\documentclass[10pt,twocolumn,letterpaper]{article}

\usepackage{wacv}
\usepackage{times}
\usepackage{epsfig}
\usepackage{graphicx}
\usepackage{amsmath}
\usepackage{amssymb}
\usepackage{multicol}
\usepackage{multirow}
\usepackage{bm} 
\usepackage{booktabs}
\usepackage{adjustbox}

\wacvfinalcopy 
\ifwacvfinal
\def\assignedStartPage{1} 
\fi

\ifwacvfinal
\usepackage[breaklinks=true,bookmarks=false]{hyperref}
\else
\usepackage[pagebackref=true,breaklinks=true,colorlinks,bookmarks=false]{hyperref}
\fi

\ifwacvfinal
\else
\fi
\newcommand{\nop}[1]{}

\begin{document}

\title{Meta Module Network for Compositional Visual Reasoning}

\author{Wenhu Chen$^2$, Zhe Gan$^1$, Linjie Li$^1$, Yu Cheng$^1$, William Wang$^2$, Jingjing Liu$^1$\\
$^1$Microsoft Dynamics 365 AI Research, Bellevue, WA, USA\\
$^2$University of California, Santa Barbara, CA, USA\\
\tt{\small\{wenhuchen,william\}@cs.ucsb.edu}\\
\tt{\small\{zhe.gan,lindsey.li,yu.cheng,jingjl\}@microsoft.com}\\
}

\maketitle

\begin{abstract}
Neural Module Network (NMN) exhibits strong interpretability and compositionality thanks to its handcrafted neural modules with explicit multi-hop reasoning capability. However, most NMNs suffer from two critical drawbacks: 1) scalability: customized module for specific function renders it impractical when scaling up to a larger set of functions in complex tasks; 2) generalizability: rigid pre-defined module inventory makes it difficult to generalize to unseen functions in new tasks/domains. To design a more powerful NMN architecture for practical use, we propose Meta Module Network (MMN) centered on a novel meta module, which can take in function recipes and morph into diverse instance modules dynamically. The instance modules are then woven into an execution graph for complex visual reasoning, inheriting the strong explainability and compositionality of NMN. With such a flexible instantiation mechanism, the parameters of instance modules are inherited from the central meta module, retaining the same model complexity as the function set grows, which promises better scalability. Meanwhile, as functions are encoded into the embedding space, unseen functions can be readily represented based on its structural similarity with previously observed ones, which ensures better generalizability. Experiments on GQA and CLEVR  datasets validate the superiority of MMN over state-of-the-art NMN designs. Synthetic experiments on held-out unseen functions from GQA dataset also demonstrate the strong generalizability of MMN. Our code and model are released in Github\footnote{\url{https://github.com/wenhuchen/Meta-Module-Network}}. 
\end{abstract}

\begin{figure}[t!]
    \centering
    \includegraphics[width=0.80\linewidth]{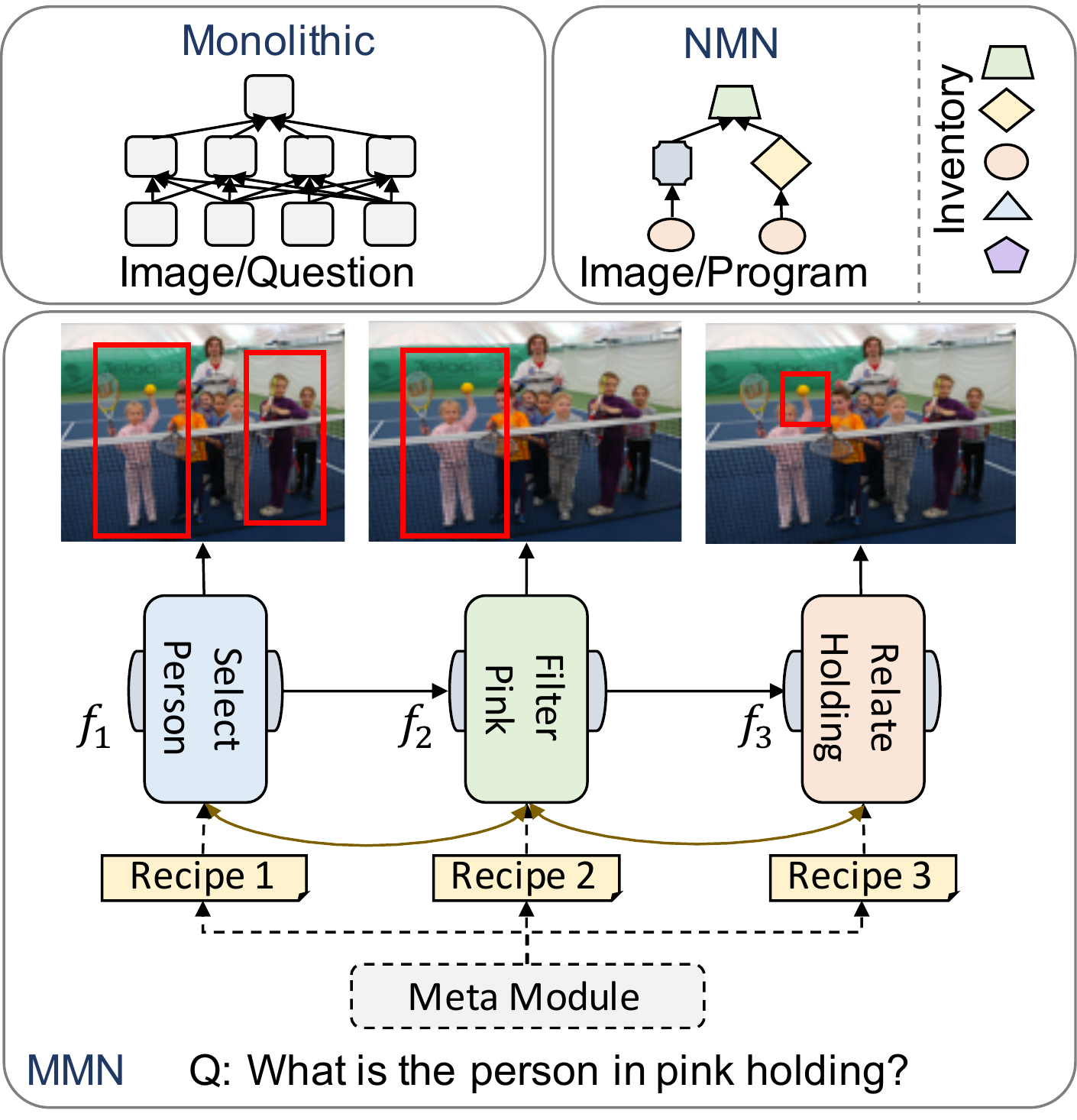}
    \caption{Comparison between NMN and MMN for visual reasoning. Neural Module Network (NMN) builds an instance-specific network based on the given question from a pre-defined inventory of neural modules, each module has its independent parameterization. Meta Module Network (MMN) also builds an instance-specific network by instantiating instance modules from the meta module based on the input function recipes (specifications), every instance module has shared parameterization.}
    \label{fig:top}
    \vspace{-3mm}
\end{figure}

\section{Introduction}
Visual reasoning requires a model to learn strong compositionality and generalization abilities, \emph{i.e.}, understanding and answering compositional questions without having seen similar semantic compositions before. Such compositional visual reasoning is a hallmark for human intelligence that endows people with strong problem-solving skills given limited prior knowledge. Neural module networks (NMNs)~\cite{andreas2016learning,andreas2016neural,hu2017learning,johnson2017inferring,hu2018explainable,mao2019neuro} have been proposed to perform such complex reasoning tasks. NMN requires a set of pre-defined functions and explicitly encodes each function into unique shallow neural networks called modules, which are composed dynamically to build an instance-specific network for each input question. This approach has high compositionality and interpretability, as each module is designed to accomplish a specific sub-task, and multiple modules can be combined to perform an unseen combination of functions during inference.

However, NMN suffers from two major limitations. 1) \textbf{Scalability}: When the complexity of the task increases, the set of functional semantics scales up, so does the number of neural modules. For example, in the recent GQA dataset~\cite{hudson2018compositional}, a larger set of functions (48 vs 25, see Appendix for details) with varied arity is involved, compared to previous CLEVR dataset~\cite{johnson2017clevr}. To solve this task with standard NMN framework~\cite{andreas2016learning,hu2017learning,mao2019neuro}, an increased amount of modules are required to implement for these functions, leading to higher model complexity. 2) \textbf{Generalizability}: Since the model is tied to a pre-defined set of functionalities when a new question with unseen functional semantics appears, no existing module can be readily applied to the new semantics, limiting the model's ability to generalize. 

In order to enhance NMN for more practical use, we propose Meta Module Network (MMN). As depicted in Figure~\ref{fig:top}, MMN is based on a meta module (a general-purpose neural network), which can take a function recipe (key-value pairs) as input to embed it into continuous vector space and feed it as a side input to instantiate different instance modules. Depending on the specification provided in function recipe, different instance modules are created to accomplish different sub-tasks. As different instance modules inherit the same parameters from the meta module, model complexity remains the same as the function set enlarges. For example, if the recipe has $K$ slots, and each slot takes $N$ values, a compact vector of the recipe can represent up to $N^K$ different functions. This effectively solves the \textbf{scalability} issue of NMN. When creating instance modules for specified sub-tasks, the input recipes are encoded into the embedding space for function instantiation. Thus, when an unseen recipe appears, it can be encoded into the embedding space to instantiate a novel instance module based on embedding similarity with previously observed recipes. This metamorphous design effectively overcomes NMN's limitation on \textbf{generalizability}. 

MMN draws inspiration from Meta Learning~\cite{schmidhuber1995learning,ravi2017optimization,finn2017model} as a learning-to-learn approach - instead of learning independent functions to solve different sub-tasks, MMN learns a meta-function that can generate a function to solve specific sub-task.\footnote{borrowing the concept of \emph{Meta} as in: a book in which a character is writing a book, or a movie in which a character is making a movie, can be described as \emph{Meta}.} The learning algorithm of MMN is based on a teacher-student framework to provide module supervision: an accurate ``symbolic teacher" first traverses a given scene graph to generate the intermediate outputs for the given functions from specific recipe; the intermediate outputs are then used as guidelines to teach each ``student" instance module to accomplish its designated sub-task in the function recipe. The module supervision together with the original question answering supervision are used jointly to train the model.

\begin{figure*}[t!]
    \centering
    \includegraphics[width=0.85\linewidth]{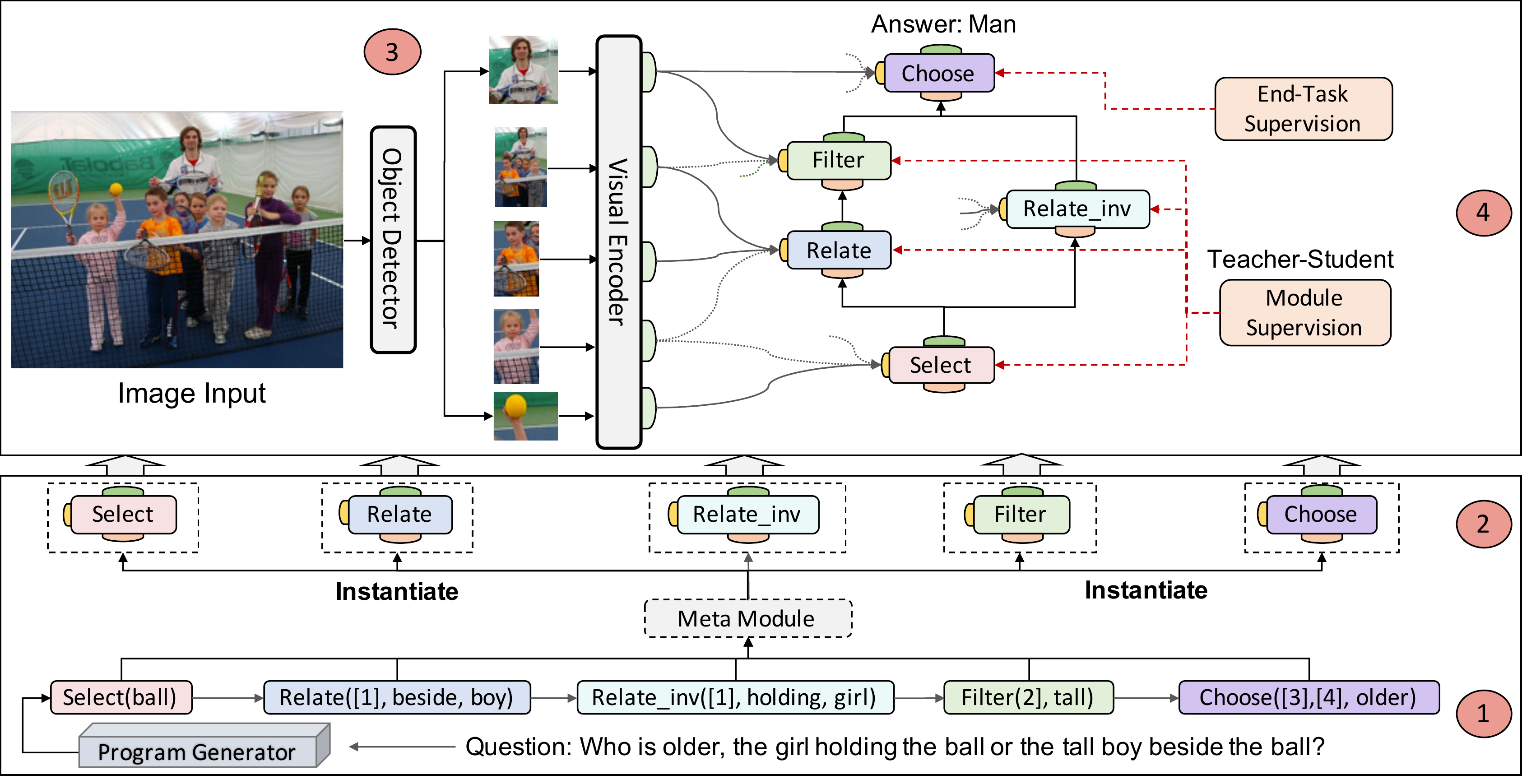}
    \caption{The model architecture of MMN: the lower part describes how the question is translated into programs and instantiated into operation-specific modules; the upper part describes the module execution. Circle $i$ denotes the $i$-th step.}
    \label{fig:architecture}
\end{figure*}

The model architecture of MMN is illustrated in Figure~\ref{fig:architecture}: ($1$) the coarse-to-fine semantic parser converts an input question into its corresponding program (\emph{i.e.}, a sequence of functions); ($2$) the meta module is instantiated into different instance modules based on the function recipes of the predicted program, which is composed into an execution graph; ($3$) the visual encoder encodes the image features that are fed to the instance modules; ($4$) during training, we provide intermediate module supervision and end-step answer supervision to jointly train all the components.

Our main contributions are summarized as follows. ($i$) We propose Meta Module Network that effectively extends the scalability and generalizability of NMN for more practical use, allowing it to handle tasks with unseen compositional function from new domain. With a metamorphous meta module learned through teacher-student supervision, MMN provides great flexibility on model design and model training that cleverly overcomes the rigid hand-crafting of NMN. ($ii$) Experiments conducted on CLEVR and GQA benchmarks demonstrate the scalability of MMN to accommodate larger set of functions. ($iii$) Qualitative visualization on the inferential chain of MMN also demonstrates its superb interpretability and strong transferability.

\section{Related Work}
\noindent\textbf{Neural Module Networks} \,
By parsing a question into a program and executing the program through dynamically composed neural modules, NMN excels in interpretability and compositionality by design~\cite{andreas2016learning,andreas2016neural,hu2017learning,johnson2017inferring,hu2018explainable,yi2018neural,mao2019neuro,vedantam2019probabilistic}. For example, IEP~\cite{hu2017learning} and N2NMN~\cite{johnson2017inferring} aims to make the whole model end-to-end trainable via the use of reinforcement learning. Stack-NMN~\cite{hu2018explainable} proposes to make soft layout selection so that the whole model is fully differentiable, and Neural-Symbolic VQA~\cite{yi2018neural} proposes to perform completely symbolic reasoning by encoding images into scene graphs.  
However, its success is mostly restricted to simple datasets with a limited set of functions, whose performance can be surpassed by simpler methods such as relational network~\cite{santoro2017simple} and FiLM~\cite{perez2018film}. Our MMN is a module network in concept, thus possessing high interpretability and compositionality. However, different from traditional NMN, to enhance its scalability and generalizability, MMN uses only a general-purpose meta module for program execution recurrently, which makes MMN inherently a monolithic network, ensuring its strong empirical performance without sacrificing model interpretability.

\vspace{5pt}
\noindent\textbf{Monolithic Network} \,
Another line of research on visual reasoning is focused on designing monolithic network architecture, such as MFB~\cite{yu2017multi}, BAN~\cite{kim2018bilinear}, DCN~\cite{nguyen2018improved}, and MCAN~\cite{yu2019deep}. These black-box models have achieved strong performance on challenging datasets, such as VQA~\cite{antol2015vqa,goyal2017making} and GQA~\cite{hudson2019gqa}, surpassing the NMN approach. More recently, multimodal pre-training algorithms~\cite{tan2019lxmert,lu2019vilbert,su2019vl,chen2019uniter,gan2020large} have been proposed that further lift state of the art on diverse tasks such as VQA~\cite{goyal2017making}, NLVR$^2$~\cite{suhr2019corpus}, and VCR~\cite{zellers2019recognition}. They use a unified neural network to learn general-purpose reasoning skills~\cite{hudson2018compositional}, which is more flexible and scalable than NMN. Most monolithic networks for visual reasoning resort to attention mechanism for multimodal fusion~\cite{zhu2017structured,zhu2016visual7w,zhou2017more,yu2019deep,yu2017multi,kim2016hadamard,kim2018bilinear,kafle2018dvqa,li2019relation,hu2019language}. To realize multi-hop reasoning on complex questions, SAN~\cite{yang2016stacked},
MAC~\cite{hudson2018compositional} and MuRel~\cite{cadene2019murel} models have been proposed. As the monolithic network is not tied to any pre-defined functionality, it has better generalizability to unseen questions. However, since the reasoning procedure is conducted in the feature space, such models usually lack interpretability, or the ability to capture the compositionality in language.

\vspace{5pt}
\noindent  \textbf{GQA Models}\,
GQA was introduced in~\cite{hudson2019gqa} for real-world visual reasoning. Simple monolithic networks~\cite{wu2019deep}, MAC netowrk~\cite{hudson2018compositional}, and language-conditioned graph neural networks~\cite{hu2019language,guo2019graph} have been developed for this task. LXMERT~\cite{tan2019lxmert}, a large-scale pre-trained encoder, has also been tested on this dataset. Recently, Neural State Machine (NSM)~\cite{hudson2019learning} proposed to first predict a probabilistic scene graph, then perform multi-hop reasoning over the graph for answer prediction. The scene graph serves as a strong prior to the model. Our model is designed to leverage dense visual features extracted from object detection models, thus orthogonal to NSM and can be enhanced with their scene graph generator once it is publicly available. Different from the aforementioned approaches, MMN also performs explicit multi-hop reasoning based on predicted programs to demonstrate inferred reasoning chain.

\section{Proposed Approach}
The visual reasoning task~\cite{hudson2019gqa} is formulated as follows: given a question $Q$ grounded in an image $I$, where $Q=\{q_1, \cdots, q_M\}$ with $q_i$ representing the $i$-th word, the goal is to select an answer $a \in \mathbb{A}$ from a set of possible answers. During training, we are provided with an additional scene graph $G$ for each image $I$, and a functional program $P$ for each question $Q$. During inference, scene graphs and programs are not provided.

Figure \ref{fig:architecture} provides an overview of Meta Module Network (MMN), which consists of three components:
($i$) Program Generator (Sec.~\ref{sec:program}), which generates a functional program from the input question;
($ii$) Visual Encoder (Sec.~\ref{sec:encoder}), which consists of self-attention and cross-attention layers on top of an object detection model, transforming an input image into object-level feature vectors; 
($iii$) Meta Module (Sec.~\ref{sec:module}), which can be instantiated to different instance modules to execute the program for answer prediction. 

\begin{figure*}[t!]
    \centering
    \includegraphics[width=0.8\linewidth]{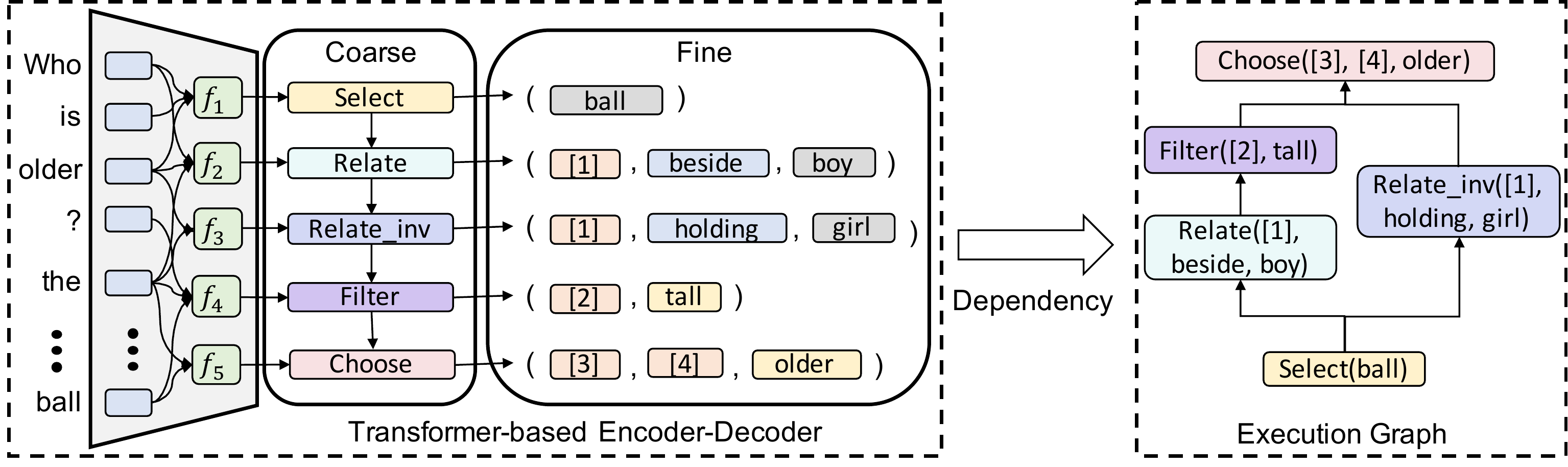}
    \caption{Architecture of the coarse-to-fine Program Generator: the left part depicts the coarse-to-fine two-stage generation; the right part depicts the resulting execution graph based on the dependency relationship.}
    \label{fig:program}
\end{figure*}
\subsection{Program Generator}
\label{sec:program}
Similar to other programming languages, we define a set of syntax rules for building valid programs and a set of semantics to determine the functionality of each program. Specifically, we define a set of functions $\mathcal{F}$ with their fixed arity $n_f \in \{1, 2, 3, 4\}$ based on the ``semantic string'' provided in GQA dataset~\cite{hudson2019gqa}. The definitions for all the functions are provided in the Appendix. The defined functions can be divided into 10 different function types (\emph{e.g.}, ``\emph{relate}'', ``\emph{verify}'', ``\emph{filter}'', ``\emph{choose}''), and each abstract function type is further implemented with different realizations based on fine-grained functionality (\emph{e.g.}, ``\emph{verify}'': ``\emph{verify\_attribute}'', ``\emph{verify\_geometric}'', ``\emph{verify\_relation}'', ``\emph{verify\_color}''), which take different arguments as inputs.  

In total, there are 48 different functions defined in GQA environment, which poses great challenges to the scalability in visual reasoning. The returned values of these functions are \emph{List of Objects}, \emph{Boolean}, or \emph{String} (\emph{Object} refers to the detected bounding box, and \emph{String} refers to object name, attributes, relations, etc.) A program $P$ is viewed as a sequence of function calls $f_1, \cdots, f_L$. For example, in Figure~\ref{fig:program}, $f_2$ is \texttt{Relate([1], beside, boy)}, the functionality of which is to find a boy who is beside the objects returned by $f_1:$ \texttt{Select(ball)}. Formally, we call $\texttt{Relate}$ the ``function name", \texttt{[1]} the ``dependency" (previous execution results), and \texttt{beside, boy} the ``arguments". By exploiting the dependency relationship between functions, we build an execution graph, where each node represents a function and each edge denotes an input-output dependency relationship between connected nodes.

In order to generate syntactically plausible programs, we follow~\cite{dong2018coarse} and adopt a coarse-to-fine two-stage generation paradigm, as illustrated in Figure~\ref{fig:program}. We first encode the question as a context vector, and then decode a sketch step by step (the sketch only contains the function name without arguments). Once the sketch is decoded, the arity and types of the decoded functions are determined. For example, after generating ``Relate", there are three arguments following this function with the first argument as the dependency. The sketch is thus expanded as ``Relate (\#1, \#2, \#3)", where ``\#$i$" denotes the $i$-th unfilled slot. We then apply a fine-grained generator to fill in the slots of dependencies and arguments for the sketch as a concrete program $P$. During the slot-filling phase, we mask the infeasible tokens at each time step to greatly reduce the search space. 

Such a two-stage generation process helps guarantee the plausibility and grammaticality of synthesized programs. For example, if function \texttt{Filter}  is sketched, we know there are two tokens required to complete the function. The first token should be selected from the dependency set (\texttt{[1]}, \texttt{[2]}, ...), while the second token should be selected from the attribute set (\emph{e.g.}, \texttt{color}, \texttt{size}). With these syntactic constraints to shrink the search space, our program synthesizer can achieve a 98.8\% execution accuracy (\emph{i.e.}, returning the same result as the ground truth after execution) compared to execution accuracy of 93\% of a standard sequence generation model.

\subsection{Visual Encoder}
\label{sec:encoder}
The visual encoder is based on a pre-trained object detection model~\cite{ren2015faster,anderson2018bottom} that extracts from image $I$ a set of regional features $\mathbf{R} = \{\mathbf{r}_i\}_{i=1}^N$, where $\mathbf{r}_i \in \mathbb{R}^{D_v}$, $N$ denotes the number of regions of interest, and $D_v$ denotes the feature dimension. Similar to a Transformer block~\cite{vaswani2017attention}, we first use two self-attention networks, $SA_{q}$ and $SA_{r}$, to encode the question and the visual features as $\hat{\mathbf{Q}} = SA_{q}(Q, Q; \phi_q)$ and $\hat{\mathbf{R}} = SA_{r}(\mathbf{R}, \mathbf{R}; \phi_r)$, respectively. $\hat{\mathbf{Q}} \in \mathbb{R}^{M \times D}$, $\hat{\mathbf{R}} \in \mathbb{R}^{N \times D}$, and $D$ is the network's hidden dimension. Based on this, a cross-attention network $CA$ is applied to use the question as guidance to refine the visual features into $\mathbf{V} = CA(\hat{\mathbf{R}}, \hat{\mathbf{Q}}; \phi_c) \in \mathbb{R}^{N \times D}$, where $\hat{\mathbf{Q}}$ is used as the query vector, and $\phi = \{\phi_q, \phi_r, \phi_c\}$ denotes all the parameters in the visual encoder. The attended visual features $\mathbf{V}$ will then be fed as the visual input to the meta module, which is detailed in Sec.~\ref{sec:module}. 

\begin{figure*}[t!]
    \centering
    \includegraphics[width=0.95\linewidth]{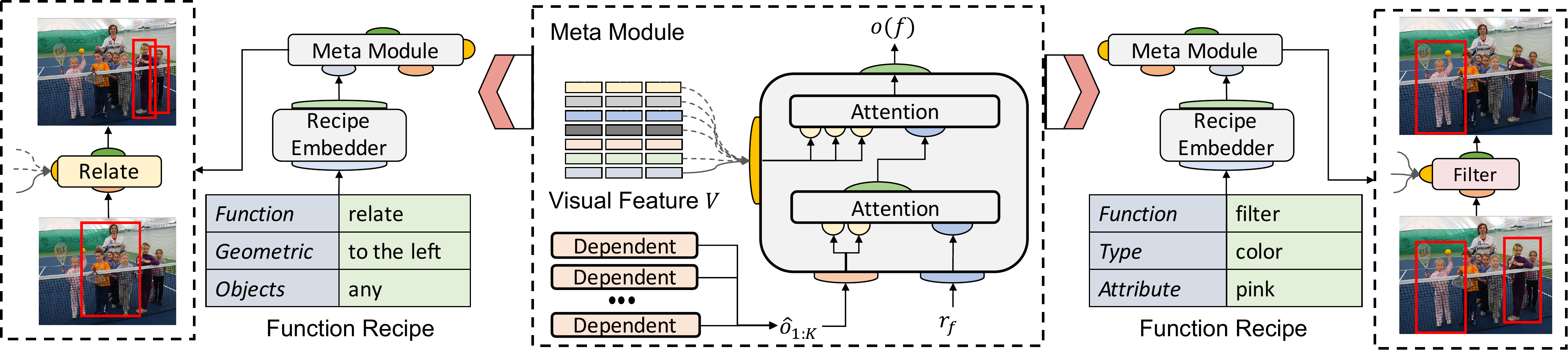}
    \caption{Illustration of the instantiation process for \texttt{Relate} and \texttt{Filter} functions.}
    \label{fig:meta}
\end{figure*}
\subsection{Meta Module}
\label{sec:module}
As opposed to having a full inventory of task-specific parameterized modules for different functions as in NMN~\cite{andreas2016neural}, we design an abstract meta module that can be instantiated into instance modules based on an input function recipe, which is a set of pre-defined key-value pairs specifying the properties of the function. As exemplified in Figure~\ref{fig:meta}, when taking recipe \texttt{Function:relate; Geometric:to the left} as the input, the Recipe Embedder produces a recipe vector to instantiate the abstract meta module into a ``geometric relation" module, which specifically searches for target objects that the current object is to the left of. When taking recipe \texttt{Function:filter; Type:color; Attribute:pink} as input, the Embedder will instantiate the meta module into a ``filter pink" module, which specifically looks for the objects with pink color in the input objects.
\begin{figure*}[t!]
    \centering
    \includegraphics[width=0.75\linewidth]{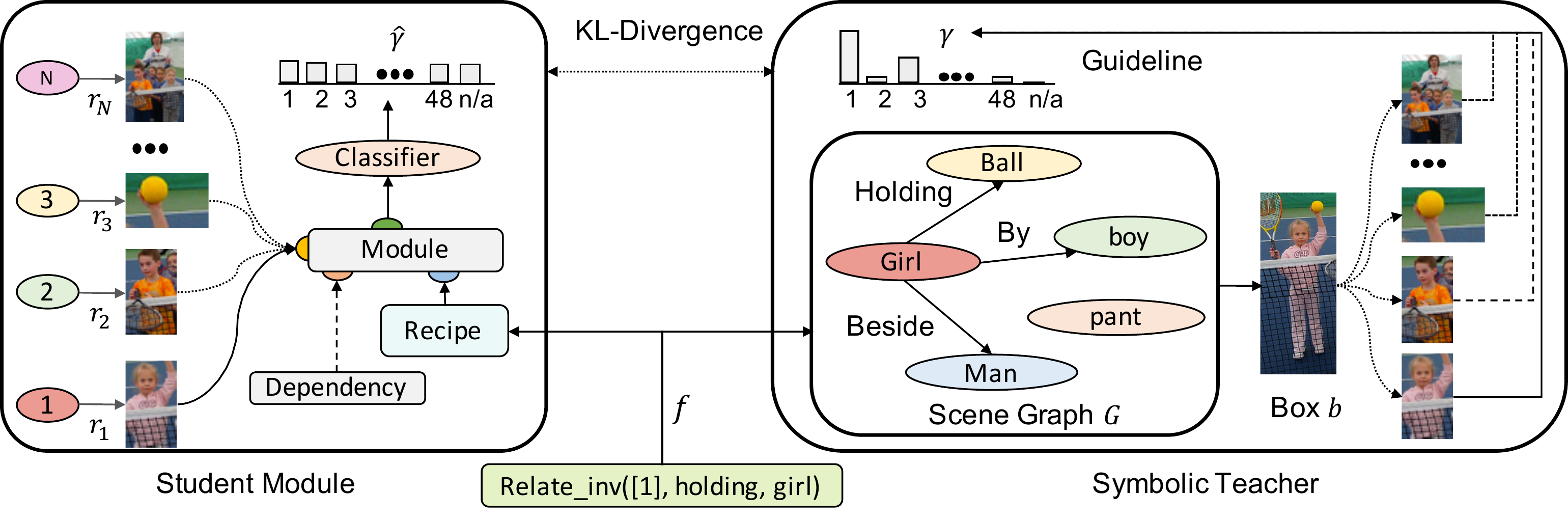}
    \caption{Illustration of the Module Supervision process: the symbolic teacher executes the function on the scene graph to obtain the bounding box ${b}$, which is then aligned with bounding boxes from the object detection model to compute the supervision guideline.}
    \label{fig:intermediate}
\end{figure*}

\vspace{5pt}
\noindent \textbf{Two-layered Attention}\, Figure~\ref{fig:meta} demonstrates the computation flow in meta module, which is built upon two-leveled multi-head attention network~\cite{vaswani2017attention}. A Recipe Embedder encodes a function recipe into a real-valued vector $\mathbf{r}_f\in \mathbb{R}^D$. In the first attention layer,  $\mathbf{r}_f$ is fed into an attention network $g_d$ as the query vector to incorporate the output ($\hat{\mathbf{o}}_{1:K}$) of dependent modules. The intermediate output ($\mathbf{o}_d$) from this attention layer is further fed into a second attention network $g_v$ to incorporate the visual representation $\mathbf{V}$ of the image. The final output is denoted as $g(\mathbf{r}_f, \hat{\mathbf{o}}_{1:K}, \mathbf{V}) = g_v(g_d(\mathbf{r}_f, \hat{\mathbf{o}}_{1:K}), \mathbf{V})$. 

\vspace{5pt}
\noindent \textbf{Instantiation \& Execution}\,
The instantiation is accomplished by feeding a function $f$ to the meta module $g$, which results in a wrapper function $g_{f}(\hat{\mathbf{o}}_{1:K}, \mathbf{V}; \psi)$ known as instance module ($\psi$ denotes the parameters of the module). Each module $g_f$ outputs $\mathbf{o}(f) \in \mathbb{R}^D$, which acts as the message passed to its neighbor modules. For brevity, we use $\mathbf{o}(f_i)$ to denote the MMN's output at the $i$-th function $f_i$. The final output $\mathbf{o}(f_L)$ of function $f_L$ will be fed into a softmax-based classifier for answer prediction. During training, we optimize the parameters $\psi$ (in meta module) and $\phi$ (in visual encoder) to maximize the likelihood $p_{\phi,\psi}(a|P, Q, \mathbf{R})$ on the training data, where $a$ is the answer, and $P, Q, \mathbf{R}$ are programs, questions and visual features.

\subsection{Learning}\label{sec:supervision}
In order to train the meta module to learn the instantiation process from given function recipes (\emph{i.e.}, how to generate functions), we propose a Teacher-Student framework depicted in~\autoref{fig:intermediate}.  First, we define a Symbolic Executor as the ``Teacher'', which can take the input function $f$ and traverse the provided training scene graph to obtain intermediate results (\emph{i.e.}, distribution over the objects on the ground-truth scene graph). The ``Teacher'' exhibits it as a guideline $\bm{\gamma}$ for the ``Student'' instance module $g_f$ to follow.

\vspace{5pt}
\noindent \textbf{Symbolic Teacher}\, We first execute the program $P=f_1, \cdots, f_L$ on the ground-truth scene graph $G$ provided in the training data to obtain all the intermediate execution results. According to the function definition (see Appendix for details), the intermediate results are either of type \emph{List of Objects} or \emph{Boolean}. The strategy of representing the results follows: $(i)$ Non-empty \emph{List of Objects}: use the first element's vertexes $[x_1, y_1, x_2, y_2]$; $(ii)$ Empty \emph{List of Objects}: use dummy vertexes $[0, 0, 0, 0]$; $(iii)$ ``True" from \emph{Boolean}: use the vertexes from last step; $(iv)$ ``False" from \emph{Boolean}: use dummy vertexes as in $(ii)$. Therefore, the intermediate results can be unified in the form of quadruples denoted as $b_i$. To align these quadruple $b_i$ regions with the regions $\mathbf{R}$ proposed by the object detector from the visual encoder, we compute its overlap against all the regions $r_j \in R$ as $a_{i, j} = \frac{Intersect(b_i, r_j)}{Union(b_i, r_j)}$. Based on whether there exist any overlaps, we handle the following two cases differently:
\begin{enumerate}
    \item If $\sum_{j} a_{i, j} > 0$, which means that there exists detected bounding boxes overlapping with the $b_i$, we normalize $a_{i, j}$ over $\mathbf{R}$ to obtain a distribution $\gamma_{i, j} = \frac{a_{i, j}}{\sum_j a_{i, j}}$ and append an extra 0 in the end to obtain $\bm{\gamma}_i \in \mathbb{R}^{N+1}$. 
    \item If $\sum_{j} a_{i, j} = 0$, which means no detected bounding box has overlap with $b_i$ (or $b_i=[0,0,0,0]$), we use the one-hot distribution $\bm{\gamma}_i = [0, \cdots, 0, 1] \in \mathbb{R}^{N+1}$ as the distribution. The last bit represents ``No Match".
\end{enumerate}
We call distributions $\gamma_{i,j}$ the guideline from symbolic teacher. Please refer to the rightmost part of Figure~\ref{fig:intermediate} to better understand the computation.

\vspace{5pt}
\noindent \textbf{Student Module}\,
We propose to demonstrate the guideline distributions $\gamma_{i,j}$ from the symbolic teacher for student instance modules $g_f$ to imitate. Formally, we let each instance module $g_f$ predict its guideline distribution based on its output representation $\mathbf{o}(f_i)$, denoted as $\hat{\bm{\gamma}}_i = softmax(MLP(\mathbf{o}(f_i)))$. During training, we enforce the instance module's prediction $\hat{\bm{\gamma}}_i$ to align the guideline distribution $\bm{\gamma}_i$ by minimizing their KL divergence $KL(\bm{\gamma}_i||\hat{\bm{\gamma}}_i)$. This side task is aimed to help the meta module learn the mapping from recipe embedding space $r_f \in \mathbb{R}^D$ to function space $f \in \mathbb{F}$ like a function factory rather than directly learning independent functions $f$ itself. Such a ``learning to learn'' or meta-learning paradigm gives our model the capability to generalize to unseen sub-tasks encoded in the recipe embedding space.

\vspace{5pt}
\noindent \textbf{Joint Optimization}\,
Formally, given the quadruple of $(P, Q, \mathbf{R}, a)$ and the pre-computed guideline distribution $\gamma$, we propose to add KL divergence to the standard loss function with a balancing factor $\eta$: 
\begin{align*}
\mathcal{L}(\phi, \psi) = -\log{p_{\phi, \psi}(a|P, Q, \mathbf{R})} + \eta\sum_{i=1}^{L-1} KL(\bm{\gamma}_i||\hat{\bm{\gamma}}_i)\,. 
\end{align*}
The objective jointly provides the module supervision and end-task supervision, the parameters $\phi, \psi$ of visual encoder and the module network are optimized w.r.t to it.

\section{Experiments}
In this section, we conduct the following experiments. $(i)$ We first evaluate the proposed Meta Module Network on CLEVR datast~\cite{johnson2017clevr} to preliminarily validate its effectiveness on the synthetic environment.  $(ii)$ We then evaluate on the GQA v1.1 dataset~\cite{hudson2019gqa} and compare it with state-of-the-art methods. As GQA is a more realistic testbed to demonstrate the scalability and generalizability of our model, we will focus on it throughout our experiments. $(iii)$ We provide visualization of the inferential chains and perform fine-grained error analysis based on that. $(iv)$ We design synthesized experiments to quantitatively measure our model's generalization ability towards unseen functional semantics.

\begin{table}[t!]
\small
\centering
\begin{tabular}{lcccccc}
\toprule
Model       & Cnt & Exist & \begin{tabular}[c]{@{}l@{}}Cmp\\ Num\end{tabular} & \begin{tabular}[c]{@{}l@{}}Cmp\\ Attr.\end{tabular} & \begin{tabular}[c]{@{}l@{}}Query\\ Attr.\end{tabular} & All \\ \midrule
NMN~\cite{hu2017learning}          & 68.5  & 85.7  & 84.9                                                     & 88.7                                                        & 90.0                                                      & 83.7    \\ 
IEP~\cite{johnson2017inferring} & 92.7  & 97.1  & 98.7                                                     & 98.9                                                        & 98.1                                                      & 96.9    \\ 
MAC~\cite{hudson2018compositional}          & 97.1  & 99.5  & 99.1                                                     & 99.5                                                        & 99.5                                                      & 98.9    \\ 
NS~\cite{yi2018neural}       & 99.7  & 99.9  & 99.9                                                     & 99.8                                                        & 99.8                                                      & 99.8    \\ 
NS-CL~\cite{mao2019neuro} & 98.2   & 98.8 & 99.0 & 99.1 & 99.3 & 98.9 \\
\midrule
MMN          & 98.2  & 99.6  & 99.3                                                     & 99.5                                                        & 99.4                                                      & 99.2    \\ \bottomrule
\end{tabular}
\vspace{1mm}
\caption{Comparison of MMN against the state-of-the-art models on CLEVR test set, as reported in their original papers.}
\label{tab:result_clevr}
\vspace{-2mm}
\end{table}

\subsection{Preliminary Experiments on CLEVR}
The CLEVR dataset~\cite{johnson2017clevr} consists of rendered images featuring 3D-objects of various shapes, materials, colors, and sizes, coupled with machine-generated compositional multi-step questions that measure performance on an array of challenging reasoning skills. Each question is also associated with a tree-structured functional program that was used to generate it, specifying the reasoning operations that should be performed to compute the answer. We use the standard training set containing 700K questions-program pairs to train our model and parser along with the provided scene graphs. We define a set of function $\mathcal{F}$ with arity of $n_f \in \{1, 2\}$ provided in the dataset. The definitions of all functions are provided in Appendix, and there are 25 functions in total with similar return type as GQA. We follow NS-VQA~\cite{yi2018neural} to train detectors based on MaskRCNN~\cite{he2017mask} and detect top 32 bounding boxes ranked by their confidence scores. We use a hidden dimension $D=256$ for both the visual encoder and the meta module. 

We report our experimental results on the standard test set in Table~\ref{tab:result_clevr}. We observe that MMN can outperform the standard NMN~\cite{hu2017learning} with more compact representation with meta module and scene-graph-based intermediate supervision. Except for numerical operations like Count and Compare Number, MMN can achieve similar accuracy as the state-of-the-art NS-VQA~\cite{yi2018neural} model. As  CLEVR dataset is not the focus of this paper due to its synthetic nature and limited set of semantic functions, we use it only as the preliminary study. Note that we have also attempted to re-implement the NS-VQA approach on GQA, and observe that the accuracy is very low ($\approx$ 30\% on test set). This is due to that NS-VQA performs pure symbolic reasoning, thus requiring the scene graph to be accurately generated; while generating scene graphs is challenging in GQA with open-domain real images.

\subsection{GQA Experimental Setup}
GQA Dataset~\cite{hudson2019gqa} contains 22M questions over 140K images. This full ``all-split" dataset has unbalanced answer distributions, thus, is further re-sampled into a ``balanced-split" with a more balanced answer distribution. The new split consists of 1M questions. Compared with the VQA v2.0 dataset~\cite{goyal2017making}, the questions in GQA are designed to require multi-hop reasoning to test the reasoning skills of developed models. Compared with the CLEVR dataset~\cite{johnson2017clevr}, GQA greatly increases the complexity of the semantic structure of questions, leading to a more diverse function set. The real-world images in GQA also bring in a bigger challenge in visual understanding. In GQA, around 94\% of questions need multi-hop reasoning, and 51\% questions are about the relationships between objects. Following~\cite{hudson2019gqa}, the evaluation metric used in our experiments is accuracy (including binary and open-ended).

The dimensionality of input image features $D_v$ is 2048, extracted from the bottom-up-attention model~\cite{anderson2018bottom}\footnote{\url{https://github.com/peteanderson80/bottom-up-attention}}. For each image, we keep the top 48 bounding boxes ranked by confidence score with the positional information of each bounding box in the form of [top-left-x, top-left-y, bottom-right-x, bottom-right-y], normalized by the image width and height. Both the meta module and the visual encoder have a hidden dimension $D$ of 512 with 8 heads. GloVe embeddings~\cite{pennington2014glove} are used to encode both questions and function keywords with 300 dimensions. The total vocabulary size is 3761, including all the functions, objects, and attributes. For training, we first use the 22M unbalanced ``all-split" to bootstrap our model with a batch size  of 2048 for 5 epochs, then fine-tune on the ``balanced-split" with a batch size of 256. The testdev-balanced split is used for model selection. 
\subsection{GQA Experimental Results}
We report our experimental results on the test2019 split (from the public GQA leaderboard) in Table~\ref{tab:results}. First, we observe significant performance gain from MMN over NMN~\cite{andreas2016neural}, which demonstrates the effectiveness of the proposed meta module mechanism. Further, we observe that our model outperforms the VQA state-of-the-art monolithic model MCAN~\cite{yu2019deep} by a large margin, which demonstrates the strong compositionality of our module-based approach. Overall, our single model achieves competitive performance (top 2) among published approaches. Notably, we achieve higher performance than LXMERT~\cite{tan2019lxmert}, which is pre-trained on large-scale out-of-domain datasets. The performance gap with NSM~\cite{hudson2019learning} is debatable since our model is standalone without relying on well-tuned external scene graph generation model~\cite{xu2017scene,yang2016stacked,chen2019knowledge}.

\begin{table}[t!]
\small
\centering
\begin{adjustbox}{scale=0.95,center}
\begin{tabular}{llccc} 
\toprule
Model        & Required Inputs & Binary & Open  & Accuracy  \\ 
\midrule
Bottom-up~\cite{anderson2018bottom} &  V+L   & 66.64  & 34.83 & 49.74     \\
MAC~\cite{hudson2018compositional}   & V+L       & 71.23  & 38.91 & 54.06     \\
GRN~\cite{guo2019graph}          & V+L  & 74.93  & 41.24 & 57.04     \\
LCGN~\cite{hu2019language}        & V+L  & 73.77  & 42.33 & 57.07     \\
BAN~\cite{kim2018bilinear} & V+L & 76.00  & 40.41 & 57.10     \\
PVR~\cite{li2019perceptual} & V+L+Program & 74.58 & 42.10 & 57.33 \\
LXMERT~\cite{tan2019lxmert} & V+L+Pre-training & 77.16  & 45.47 & 60.33     \\ 
NSM~\cite{hudson2019learning} & V+L+SceneModel & \textbf{78.94}  & \textbf{49.25} & \textbf{63.17}     \\ 
\midrule
MCAN~\cite{yu2019deep}          & V+L  & 75.87   & 42.15 & 57.96     \\
NMN~\cite{andreas2016neural}    & V+L+Program & 72.88  & 40.53 & 55.70\\
MMN (Ours)  & V+L+Program & \textbf{78.90}  & \textbf{44.89} & \textbf{60.83}     \\
\bottomrule
\end{tabular}
\end{adjustbox}
\vspace{1mm}
\caption{Comparison of MMN single model with published state-of-the-art methods on the blind test2019 leaderboard.}
\label{tab:results}
\end{table}
To verify the contribution of each component in MMN, we perform several ablation studies. (1) \emph{w/o Module Supervision vs. w/ Module Supervision}. We investigate the influence of module supervision by changing the hyper-parameter $\eta$ from $0$ to $2.0$ to see how much influence the module supervision has on the model performance.
(2) \emph{w/o Bootstrap vs. w/ Bootstrap}. We investigate the effectiveness of bootstrapping in training to validate whether we could use the large-scale unbalanced split to benefit on the model's performance. 

\begin{table}[t!]
\small
\centering
\begin{tabular}{lc|lc} 
\toprule
Ablation (1)            & Accuracy &  Ablation (2)       & Accuracy  \\ 
\midrule
MCAN & 57.4                & MMN w/o BS      & 58.4      \\
MMN ($\eta$ = 0)  & 58.1                  & MMN w/o FT       & 56.5      \\
MMN ($\eta$ = 0.1) & 59.1     & MMN + BS (2 ep) & 59.2      \\
MMN ($\eta$ = 0.5) & \textbf{60.4}     &  MMN + BS (3 ep) & 59.9      \\
MMN ($\eta$ = 1.0) & 60.1     & MMN + BS (4 ep) & \textbf{60.4}     \\
MMN ($\eta$ = 2.0) & 59.5     & MMN + BS (5 ep) & 60.0      \\
\bottomrule
\end{tabular}
\vspace{1mm}
\caption{Ablation study on GQA testdev. BS means bootstrapping, FT means fine-tuning; w/o BS: Directly training on the balanced-split; (n ep) means bootstrapped for n epochs. }
\label{tab:ablation_module}
\vspace{-1mm}
\end{table}

\begin{figure*}[!thb]
    \centering
    \includegraphics[width=0.80\linewidth]{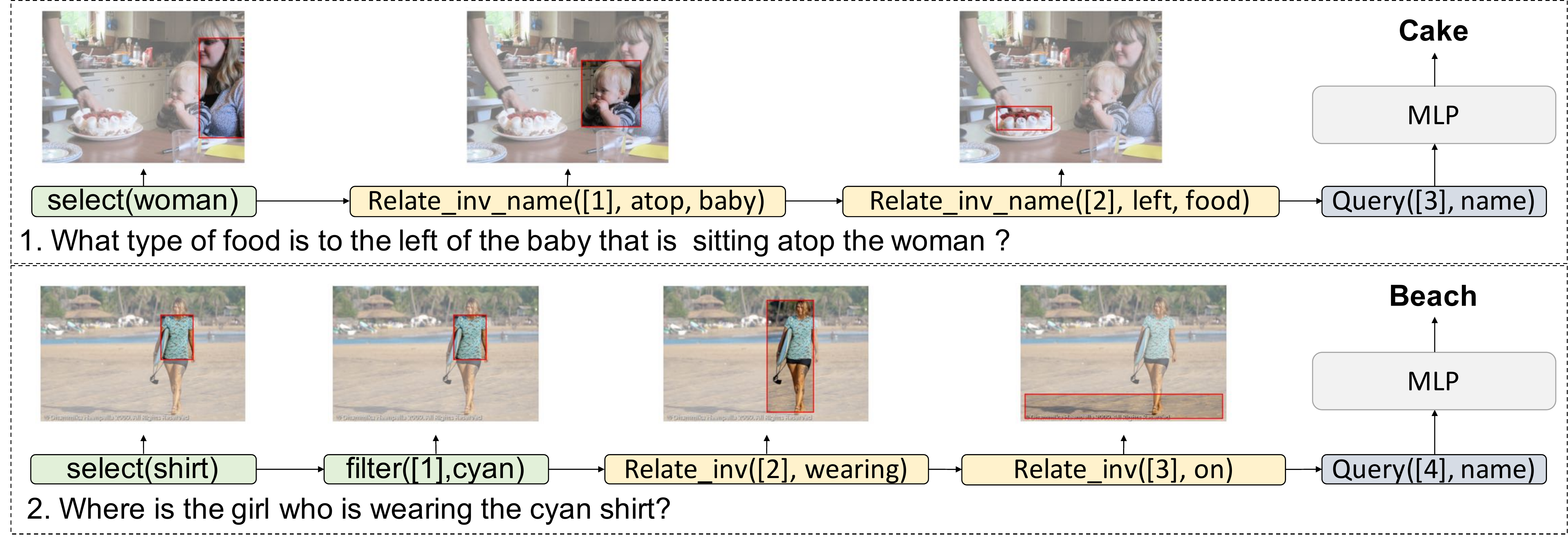}
    \vspace{-1mm}
    \caption{Visualization of the inferential chains learned by our model.}
    \label{fig:inferential_chain}
\end{figure*}

We further report the ablation results for the validation split in Table~\ref{tab:ablation_module}. From Ablation (1), we observe that without module supervision, our MMN already achieves decent improvement over 6-layered MCAN~\cite{yu2019deep}. Since all the modules have shared parameters, our model has similar parameter size as 1-layered MCAN. The result demonstrates the efficiency of the parameterization in our MMN. By increasing $\eta$ from 0.1 to 0.5, accuracy steadily improves, which reflects the effectiveness of module supervision. Further increasing the value of $\eta$ did not improve the performance empirically. 
From Ablation (2), we observe that bootstrapping is a critical step for MMN, as it explores more data to better regularize functionalities of reasoning modules. Bootstrap for 4 epochs can yield better performance in our experiments.
\begin{table*}[!t]
\small
\centering
\begin{tabular}{l|c|c|c|c|c|c} 
\hline
Function & \multicolumn{3}{c|}{verify\_shape} & \multicolumn{3}{c}{relate\_name}  \\ 
\hline
Methods  & NMN & 0\%(MMN) & 100\%(MMN)                                    & NMN & 0\%(MMN) & 100\%(MMN)  \\ 
\hline
Accuracy & 50\%     & 61\%  & 74\%                          & 5\%      & 23\%    & 49\% \\
\hline
Function & \multicolumn{3}{c|}{filter\_location} & \multicolumn{3}{c}{choose\_name}  \\ 
\hline
Methods  & NMN & 0\%(MMN) & 100\%(MMN)                           & NMN & 0\%(MMN) & 100\%(MMN)         \\ 
\hline
Accuracy & 50\%      & 77\% &  86\%                             & 50\%     & 62\% & 79\%\\
\hline
\end{tabular}
\vspace{2mm}
\caption{Analysis of MMN's generalizability to unseen functions. In NMN, since the unseen function does not have pre-defined module, the performance is close to randomness. In MMN, 0\% means without any training instances, 100\% means fully-supervision.}
\label{tab:zero_shot}
\end{table*}
\subsection{Generalization Experimental Results}
\label{sec:generalization}
Similar to Meta Learning~\cite{finn2017model}, we also evaluate whether our meta module has learned the ability to adapt to unseen sub-tasks. To evaluate such generalization ability, we perform additional experiments, where we held out all the training instances containing \texttt{verify\_shape, relate\_name}, \texttt{filter\_location, choose\_name} to quantitatively measure model's performance on these unseen functions. Standard NMN~\cite{andreas2016neural} fails to handle these unseen functions, as it requires training instances for the randomly initialized shallow module network for these unseen functions. In contrast, MMN can generalize the unseen functions from recipe space and exploits the structural similarity with its related functions to infer its semantic functionality. For example, if the training set contains \texttt{verify\_size} (function: verify, type: size, attr: ?) and \texttt{filter\_shape} (function: filter, type: shape, attr: ?) functions in the recipes, and instantiated module is capable of inferring the functionality of an unseen but similar function \texttt{verify\_shape} (function: verify, type:shape, attr: ?) from the recipe embedding space. Table~\ref{tab:zero_shot} shows that the zero-shot accuracy of the proposed meta module is significantly higher than NMN (equivalent to random guess), which demonstrates the generalizability of proposed MMN architecture. Instead of handcrafting a new module every time when new function appears like NMN~\cite{andreas2016neural}, our MMN is more flexible and extensible for handling growing function sets. Such observation further validates the value of the proposed method to adapt a more challenging environment where we need to handle unknown functions.

\subsection{Interpretability and Error Analysis}
\label{sec:error}
To demonstrate the interpretability of MMN, Figure~\ref{fig:inferential_chain} provides some visualization results to show the inferential chain during reasoning. As shown, the model correctly executes the intermediate results and yields the correct final answer. More visualization examples are provided in the Appendix. To better interpret the model's behavior, we also perform quantitative analysis to diagnose the errors in the inferential chain. Here, we held out a small validation set to analyze the execution accuracy of different functions. Our model obtains Recall@1 of 59\% and Recall@2 of 73\%, which indicates that the object selected by the symbolic teacher has 59\% chance of being top-1, and 73\% chance as the top-2 by the student model, significantly higher than random-guess Recall@1 of 2\%, demonstrating the effectiveness of module supervision. 

Furthermore, we conduct a detailed analysis of function-wise execution accuracy to understand the limitation of MMN. We found that most erroneous functions are \texttt{relate} and \texttt{query}, having 44\% and 60\% execution accuracy respectively. These errors are mainly related to scene understanding, which suggests that the scene graph model is critical to surpassing NSM~\cite{hudson2019learning} on performance. However, this is out of the scope of this paper and we plan to leave it for future study.

\section{Conclusion}
In this paper, we propose Meta Module Network that resolves the known challenges of NMN. Our model is built upon a meta module, which can be instantiated into an instance module to perform designated functionalities dynamically. Our approach can significantly outperform baseline methods and achieve comparable performance to state of the art while maintaining strong explainability. 

{\small
\bibliographystyle{ieee_fullname}
\bibliography{egbib}

\begin{thebibliography}{10}\itemsep=-1pt

\bibitem{anderson2018bottom}
Peter Anderson, Xiaodong He, Chris Buehler, Damien Teney, Mark Johnson, Stephen
  Gould, and Lei Zhang.
\newblock Bottom-up and top-down attention for image captioning and visual
  question answering.
\newblock In {\em CVPR}, 2018.

\bibitem{andreas2016learning}
Jacob Andreas, Marcus Rohrbach, Trevor Darrell, and Dan Klein.
\newblock Learning to compose neural networks for question answering.
\newblock In {\em NAACL}, 2016.

\bibitem{andreas2016neural}
Jacob Andreas, Marcus Rohrbach, Trevor Darrell, and Dan Klein.
\newblock Neural module networks.
\newblock In {\em CVPR}, 2016.

\bibitem{antol2015vqa}
Stanislaw Antol, Aishwarya Agrawal, Jiasen Lu, Margaret Mitchell, Dhruv Batra,
  C Lawrence~Zitnick, and Devi Parikh.
\newblock Vqa: Visual question answering.
\newblock In {\em ICCV}, 2015.

\bibitem{cadene2019murel}
Remi Cadene, Hedi Ben-Younes, Matthieu Cord, and Nicolas Thome.
\newblock Murel: Multimodal relational reasoning for visual question answering.
\newblock In {\em CVPR}, 2019.

\bibitem{chen2019knowledge}
Tianshui Chen, Weihao Yu, Riquan Chen, and Liang Lin.
\newblock Knowledge-embedded routing network for scene graph generation.
\newblock In {\em CVPR}, 2019.

\bibitem{chen2019uniter}
Yen-Chun Chen, Linjie Li, Licheng Yu, Ahmed~El Kholy, Faisal Ahmed, Zhe Gan, Yu
  Cheng, and Jingjing Liu.
\newblock Uniter: Learning universal image-text representations.
\newblock {\em arXiv preprint arXiv:1909.11740}, 2019.

\bibitem{dong2018coarse}
Li Dong and Mirella Lapata.
\newblock Coarse-to-fine decoding for neural semantic parsing.
\newblock In {\em ACL}, 2018.

\bibitem{finn2017model}
Chelsea Finn, Pieter Abbeel, and Sergey Levine.
\newblock Model-agnostic meta-learning for fast adaptation of deep networks.
\newblock In {\em ICML}, 2017.

\bibitem{gan2020large}
Zhe Gan, Yen-Chun Chen, Linjie Li, Chen Zhu, Yu Cheng, and Jingjing Liu.
\newblock Large-scale adversarial training for vision-and-language
  representation learning.
\newblock {\em arXiv preprint arXiv:2006.06195}, 2020.

\bibitem{goyal2017making}
Yash Goyal, Tejas Khot, Douglas Summers-Stay, Dhruv Batra, and Devi Parikh.
\newblock Making the v in vqa matter: Elevating the role of image understanding
  in visual question answering.
\newblock In {\em CVPR}, 2017.

\bibitem{guo2019graph}
Dalu Guo, Chang Xu, and Dacheng Tao.
\newblock Graph reasoning networks for visual question answering.
\newblock {\em arXiv preprint arXiv:1907.09815}, 2019.

\bibitem{he2017mask}
Kaiming He, Georgia Gkioxari, Piotr Doll{\'a}r, and Ross Girshick.
\newblock Mask r-cnn.
\newblock In {\em ICCV}, 2017.

\bibitem{hu2018explainable}
Ronghang Hu, Jacob Andreas, Trevor Darrell, and Kate Saenko.
\newblock Explainable neural computation via stack neural module networks.
\newblock In {\em ECCV}, 2018.

\bibitem{hu2017learning}
Ronghang Hu, Jacob Andreas, Marcus Rohrbach, Trevor Darrell, and Kate Saenko.
\newblock Learning to reason: End-to-end module networks for visual question
  answering.
\newblock In {\em ICCV}, 2017.

\bibitem{hu2019language}
Ronghang Hu, Anna Rohrbach, Trevor Darrell, and Kate Saenko.
\newblock Language-conditioned graph networks for relational reasoning.
\newblock In {\em ICCV}, 2019.

\bibitem{hudson2018compositional}
Drew~A Hudson and Christopher~D Manning.
\newblock Compositional attention networks for machine reasoning.
\newblock In {\em ICLR}, 2018.

\bibitem{hudson2019gqa}
Drew~A Hudson and Christopher~D Manning.
\newblock Gqa: A new dataset for real-world visual reasoning and compositional
  question answering.
\newblock In {\em CVPR}, 2019.

\bibitem{hudson2019learning}
Drew~A Hudson and Christopher~D Manning.
\newblock Learning by abstraction: The neural state machine.
\newblock In {\em NeurIPS}, 2019.

\bibitem{johnson2017clevr}
Justin Johnson, Bharath Hariharan, Laurens van~der Maaten, Li Fei-Fei, C
  Lawrence~Zitnick, and Ross Girshick.
\newblock Clevr: A diagnostic dataset for compositional language and elementary
  visual reasoning.
\newblock In {\em CVPR}, 2017.

\bibitem{johnson2017inferring}
Justin Johnson, Bharath Hariharan, Laurens van~der Maaten, Judy Hoffman, Li
  Fei-Fei, C Lawrence~Zitnick, and Ross Girshick.
\newblock Inferring and executing programs for visual reasoning.
\newblock In {\em ICCV}, 2017.

\bibitem{kafle2018dvqa}
Kushal Kafle, Brian Price, Scott Cohen, and Christopher Kanan.
\newblock Dvqa: Understanding data visualizations via question answering.
\newblock In {\em CVPR}, 2018.

\bibitem{kim2018bilinear}
Jin-Hwa Kim, Jaehyun Jun, and Byoung-Tak Zhang.
\newblock Bilinear attention networks.
\newblock In {\em NeurIPS}, 2018.

\bibitem{kim2016hadamard}
Jin-Hwa Kim, Kyoung-Woon On, Woosang Lim, Jeonghee Kim, Jung-Woo Ha, and
  Byoung-Tak Zhang.
\newblock Hadamard product for low-rank bilinear pooling.
\newblock In {\em ICLR}, 2016.

\bibitem{li2019perceptual}
Guohao Li, Xin Wang, and Wenwu Zhu.
\newblock Perceptual visual reasoning with knowledge propagation.
\newblock In {\em ACMMM}, 2019.

\bibitem{li2019relation}
Linjie Li, Zhe Gan, Yu Cheng, and Jingjing Liu.
\newblock Relation-aware graph attention network for visual question answering.
\newblock {\em arXiv preprint arXiv:1903.12314}, 2019.

\bibitem{lu2019vilbert}
Jiasen Lu, Dhruv Batra, Devi Parikh, and Stefan Lee.
\newblock Vilbert: Pretraining task-agnostic visiolinguistic representations
  for vision-and-language tasks.
\newblock In {\em NeurIPS}, 2019.

\bibitem{mao2019neuro}
Jiayuan Mao, Chuang Gan, Pushmeet Kohli, Joshua~B Tenenbaum, and Jiajun Wu.
\newblock The neuro-symbolic concept learner: Interpreting scenes, words, and
  sentences from natural supervision.
\newblock In {\em ICLR}, 2019.

\bibitem{nguyen2018improved}
Duy-Kien Nguyen and Takayuki Okatani.
\newblock Improved fusion of visual and language representations by dense
  symmetric co-attention for visual question answering.
\newblock In {\em CVPR}, 2018.

\bibitem{pennington2014glove}
Jeffrey Pennington, Richard Socher, and Christopher Manning.
\newblock Glove: Global vectors for word representation.
\newblock In {\em EMNLP}, 2014.

\bibitem{perez2018film}
Ethan Perez, Florian Strub, Harm De~Vries, Vincent Dumoulin, and Aaron
  Courville.
\newblock Film: Visual reasoning with a general conditioning layer.
\newblock In {\em AAAI}, 2018.

\bibitem{ravi2017optimization}
Sachin Ravi and Hugo Larochelle.
\newblock Optimization as a model for few-shot learning.
\newblock {\em ICLR}, 2017.

\bibitem{ren2015faster}
Shaoqing Ren, Kaiming He, Ross Girshick, and Jian Sun.
\newblock Faster r-cnn: Towards real-time object detection with region proposal
  networks.
\newblock In {\em NeurIPS}, 2015.

\bibitem{santoro2017simple}
Adam Santoro, David Raposo, David~G Barrett, Mateusz Malinowski, Razvan
  Pascanu, Peter Battaglia, and Timothy Lillicrap.
\newblock A simple neural network module for relational reasoning.
\newblock In {\em NeurIPS}, 2017.

\bibitem{schmidhuber1995learning}
J{\"u}rgen Schmidhuber.
\newblock On learning how to learn learning strategies.
\newblock {\em Citeseer}, 1995.

\bibitem{su2019vl}
Weijie Su, Xizhou Zhu, Yue Cao, Bin Li, Lewei Lu, Furu Wei, and Jifeng Dai.
\newblock Vl-bert: Pre-training of generic visual-linguistic representations.
\newblock {\em arXiv preprint arXiv:1908.08530}, 2019.

\bibitem{suhr2019corpus}
Alane Suhr, Stephanie Zhou, Ally Zhang, Iris Zhang, Huajun Bai, and Yoav Artzi.
\newblock A corpus for reasoning about natural language grounded in
  photographs.
\newblock {\em ACL}, 2019.

\bibitem{tan2019lxmert}
Hao Tan and Mohit Bansal.
\newblock Lxmert: Learning cross-modality encoder representations from
  transformers.
\newblock In {\em EMNLP}, 2019.

\bibitem{vaswani2017attention}
Ashish Vaswani, Noam Shazeer, Niki Parmar, Jakob Uszkoreit, Llion Jones,
  Aidan~N Gomez, {\L}ukasz Kaiser, and Illia Polosukhin.
\newblock Attention is all you need.
\newblock In {\em NeurIPS}, 2017.

\bibitem{vedantam2019probabilistic}
Ramakrishna Vedantam, Karan Desai, Stefan Lee, Marcus Rohrbach, Dhruv Batra,
  and Devi Parikh.
\newblock Probabilistic neural-symbolic models for interpretable visual
  question answering.
\newblock In {\em ICML}, 2019.

\bibitem{wu2019deep}
Chenfei Wu, Yanzhao Zhou, Gen Li, Nan Duan, Duyu Tang, and Xiaojie Wang.
\newblock Deep reason: A strong baseline for real-world visual reasoning.
\newblock {\em arXiv preprint arXiv:1905.10226}, 2019.

\bibitem{xu2017scene}
Danfei Xu, Yuke Zhu, Christopher~B Choy, and Li Fei-Fei.
\newblock Scene graph generation by iterative message passing.
\newblock In {\em CVPR}, 2017.

\bibitem{yang2016stacked}
Zichao Yang, Xiaodong He, Jianfeng Gao, Li Deng, and Alex Smola.
\newblock Stacked attention networks for image question answering.
\newblock In {\em CVPR}, 2016.

\bibitem{yi2018neural}
Kexin Yi, Jiajun Wu, Chuang Gan, Antonio Torralba, Pushmeet Kohli, and Josh
  Tenenbaum.
\newblock Neural-symbolic vqa: Disentangling reasoning from vision and language
  understanding.
\newblock In {\em NeurIPS}, 2018.

\bibitem{yu2019deep}
Zhou Yu, Jun Yu, Yuhao Cui, Dacheng Tao, and Qi Tian.
\newblock Deep modular co-attention networks for visual question answering.
\newblock In {\em CVPR}, 2019.

\bibitem{yu2017multi}
Zhou Yu, Jun Yu, Jianping Fan, and Dacheng Tao.
\newblock Multi-modal factorized bilinear pooling with co-attention learning
  for visual question answering.
\newblock In {\em ICCV}, 2017.

\bibitem{zellers2019recognition}
Rowan Zellers, Yonatan Bisk, Ali Farhadi, and Yejin Choi.
\newblock From recognition to cognition: Visual commonsense reasoning.
\newblock In {\em CVPR}, 2019.

\bibitem{zhou2017more}
Yiyi Zhou, Rongrong Ji, Jinsong Su, Yongjian Wu, and Yunsheng Wu.
\newblock More than an answer: Neural pivot network for visual qestion
  answering.
\newblock In {\em ACMMM}, 2017.

\bibitem{zhu2017structured}
Chen Zhu, Yanpeng Zhao, Shuaiyi Huang, Kewei Tu, and Yi Ma.
\newblock Structured attentions for visual question answering.
\newblock In {\em ICCV}, 2017.

\bibitem{zhu2016visual7w}
Yuke Zhu, Oliver Groth, Michael Bernstein, and Li Fei-Fei.
\newblock Visual7w: Grounded question answering in images.
\newblock In {\em CVPR}, 2016.

\end{thebibliography}
}

\clearpage
\onecolumn
\appendix
\section{Appendix for ``Meta Module Network for Compositional Visual Reasoning''}
\subsection{Visual Encoder and Multi-head Attention}
\begin{figure}[!h]
    \centering
    \includegraphics[width=0.85\textwidth]{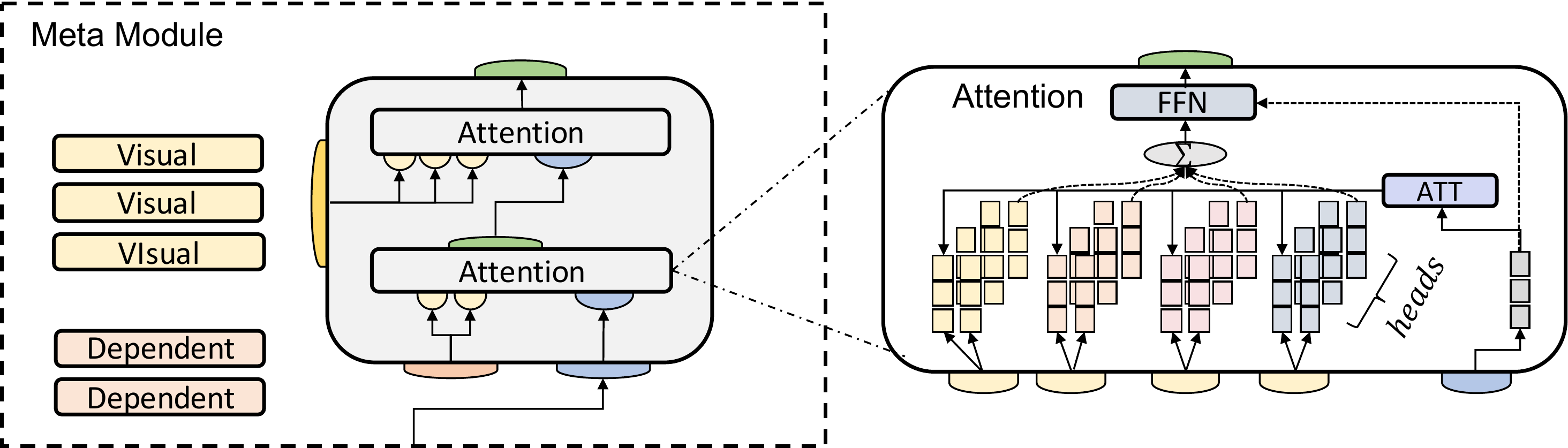}
    \caption{Illustration of the multi-head attention network used in the Meta Module.}
    \label{fig:attention}
\end{figure}

The multi-head attention network is illustrated in Figure~\ref{fig:attention}.

\subsection{Recipe Embedding}
\begin{figure}[!h]
    \centering
    \includegraphics[width=0.70\textwidth]{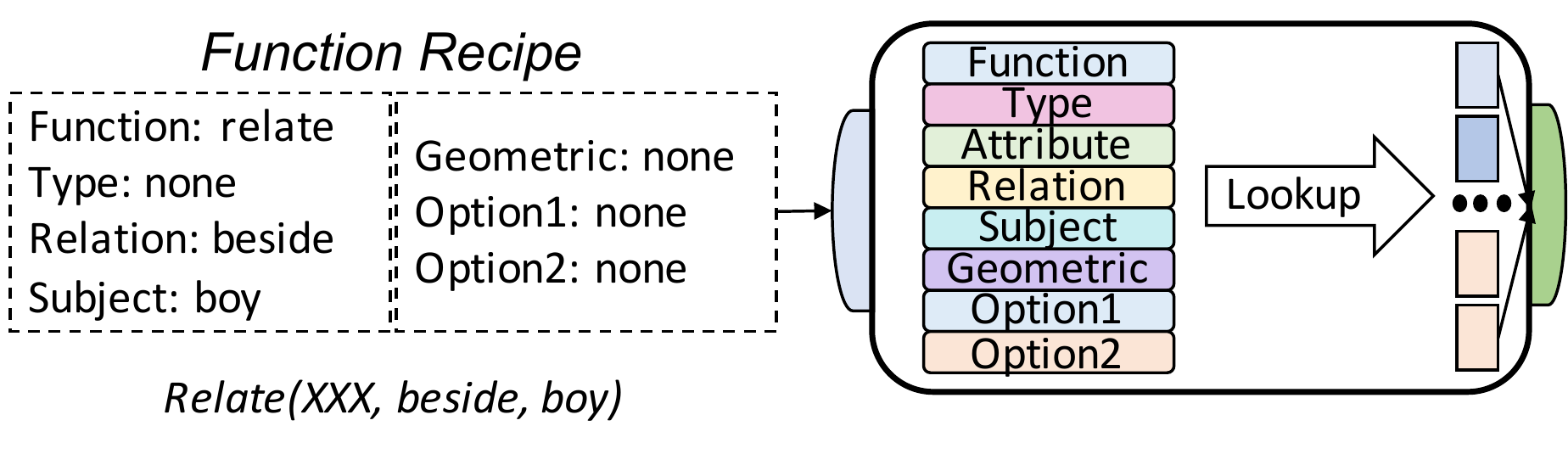}
    \caption{Illustration of the recipe embedder.}
    \label{fig:embedding}
\end{figure}

The recipe embedder is illustrated in Figure~\ref{fig:embedding}.

\subsection{Implementation}
\begin{figure}[!h]
    \centering
    \includegraphics[width=0.95\textwidth]{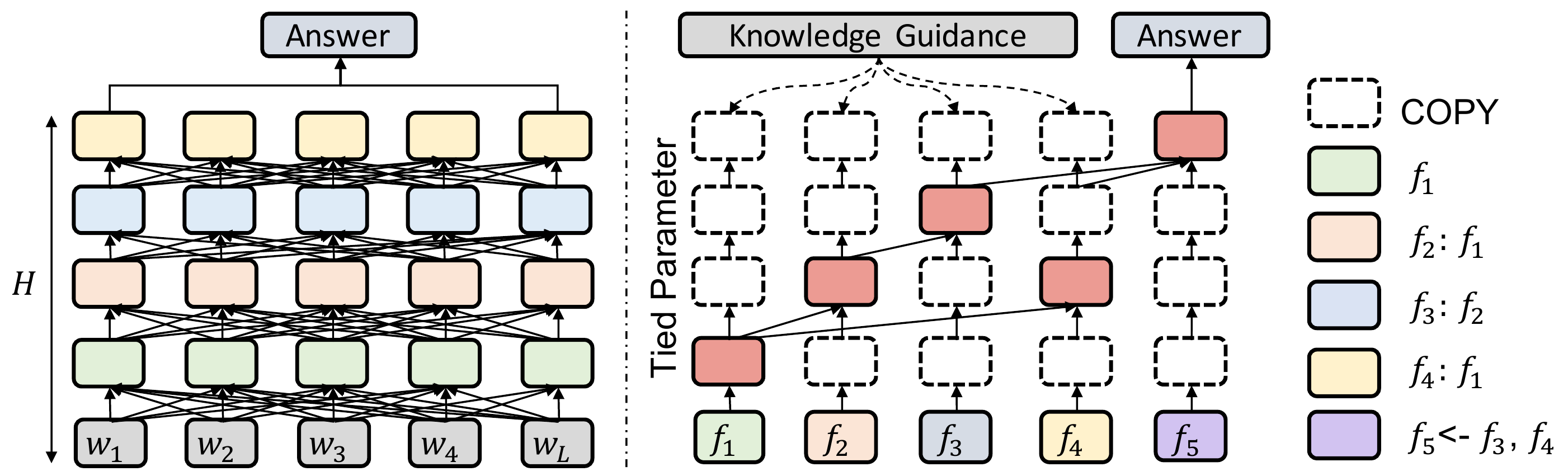}
    \caption{Illustration of the implementation of both the Transformer model (left) and our model (right).}
    \label{fig:implem}
\end{figure}

The implementation of the proposed model is demonstrated in Figure~\ref{fig:implem}. Our model can be efficiently implemented by adding masks on top of the Transformer model, guided by an additional supervision signal.

\subsection{Function Statistics}
The function statistics is listed in Table~\ref{tab:functions}.
\begin{table}[!h]
    \centering
    \small
    \begin{tabular}{lccccccccccc}
    \toprule
    \textbf{Type} & Relate &  Select &  Filter & Choose & Verify & Query & Common & Differ & Bool & Exist & All \\
    \midrule
    \textbf{Funcs} & 5 & 1 & 8 & 12 & 5 & 6 & 2 & 6 & 2 & 1 & 48 \\
    \bottomrule
    \end{tabular}
    \caption{The statistics of different functions.}
    \label{tab:functions}
\end{table}

\subsection{Inferential Chains}
\begin{figure}[htb]
    \centering
    \includegraphics[width=1.0\linewidth]{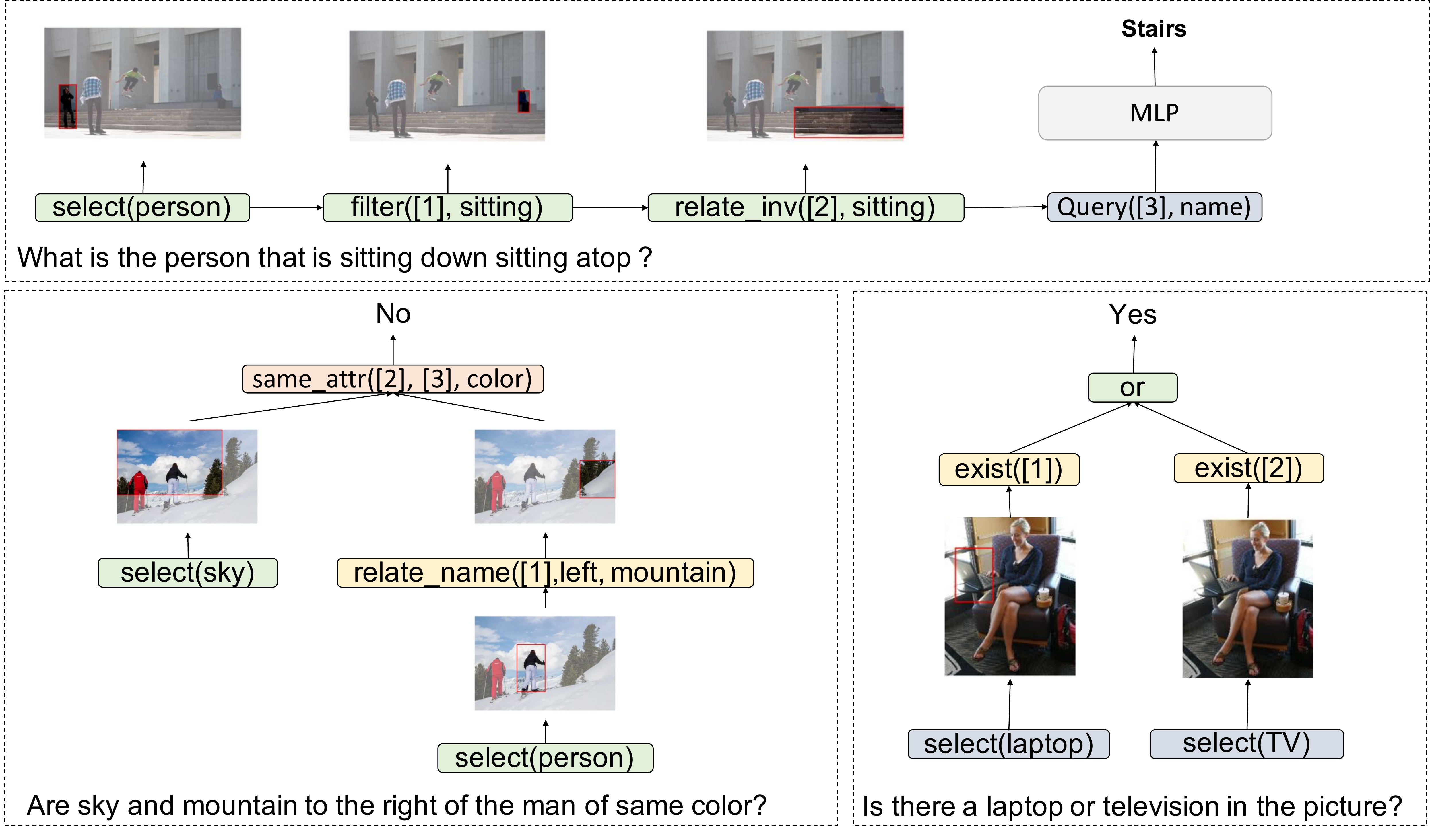}
    \caption{More examples on visualization of the inferential chains learned by our model.}
    \label{fig:vis1}
\end{figure}

\begin{figure}[htb]
    \centering
    \includegraphics[width=0.85\linewidth]{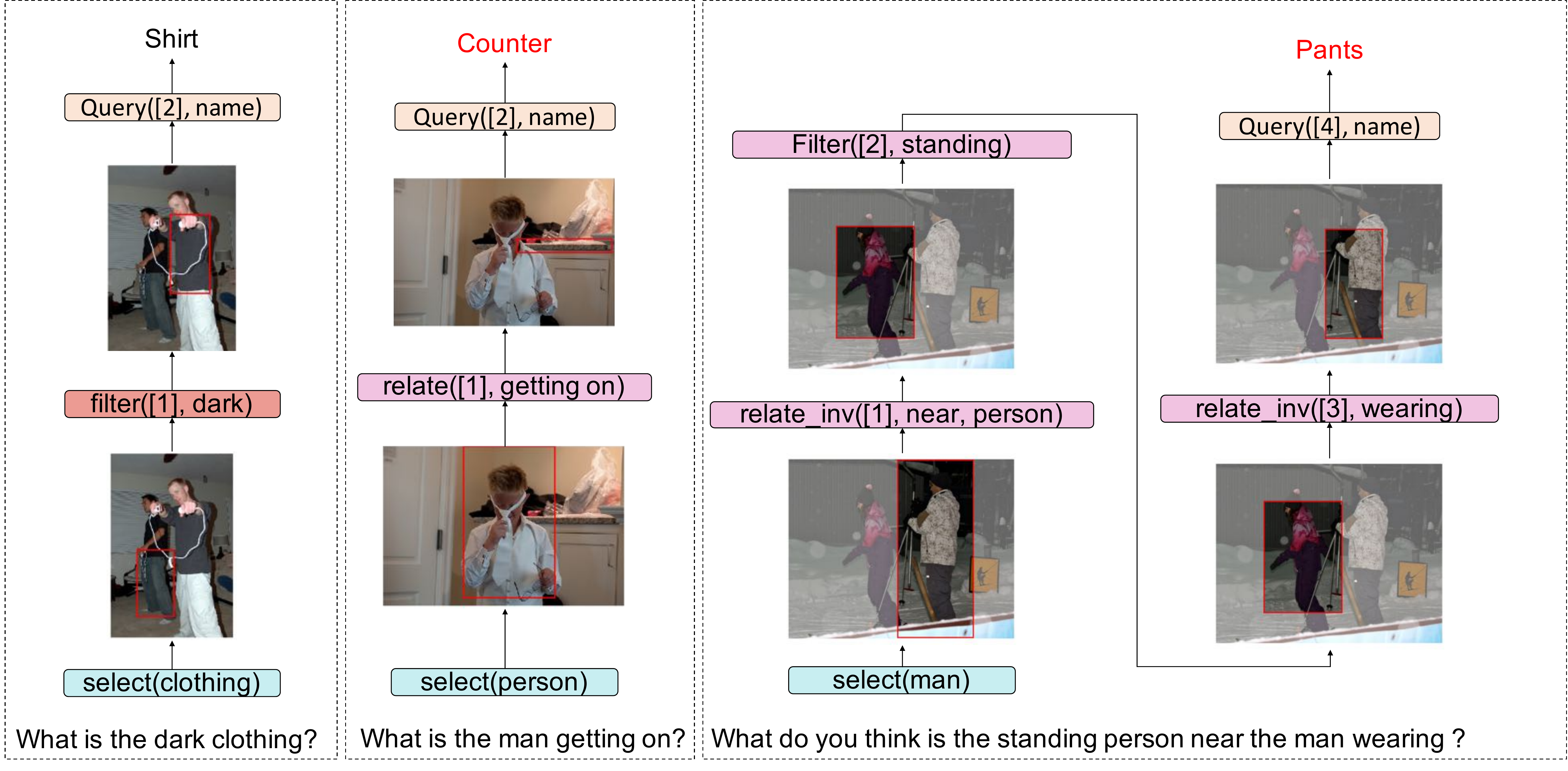}
    \caption{More examples on visualization of the inferential chains learned by our model.}
    \label{fig:vis2}
\end{figure}

More inferential chains are visualized in Figure~\ref{fig:vis1} and~\ref{fig:vis2}.

\subsection{Detailed Error Analysis}
Furthermore, we conduct a detailed analysis of function-wise execution accuracy to understand the limitation of MMN. Results are shown in Table~\ref{tab:error_analysis}. Below are the observed main bottlenecks: $(i)$ relation-type functions such as \texttt{relate, relate\_inv}; and $(ii)$ object/attribute recognition functions such as \texttt{query\_name, query\_color}. We hypothesize that this might be attributed to the quality of visual features from standard object detection models~\cite{anderson2018bottom}, which does not capture the relations between objects well. Besides, the object and attribute classification network are not fine-tuned on GQA. This suggests that scene graph modeling for visual scene understanding is critical to surpassing NSM~\cite{hudson2019learning} on performance.

\begin{table}[t!]
\small
\centering
\begin{tabular}{cccccc|ccc|cc}
\toprule
\multicolumn{6}{c}{Binary} & \multicolumn{3}{|c|}{Objects} & \multicolumn{2}{c}{String} \\
\midrule
verify  & choose & compare  & exist & and & or  & filter & select  & relate & query[object] & query[scene] \\
\midrule
0.74   & 0.79  & 0.88     &  0.88  & 0.97 & 0.95  & 0.67   & 0.61 & 0.44 & 0.61 & 0.65 \\         
\bottomrule
\end{tabular}
\vspace{1mm}
\caption{Error analysis on different types of functions. ``Objects" functions only appear in the intermediate step, ``String" function only appears in the final step, ``Binary" functions can occur in both intermediate and final step.}
\label{tab:error_analysis}
\vspace{-4mm}
\end{table}

\subsection{Function Description}
\begin{figure}[!h]
    \centering
    \includegraphics[width=0.9\linewidth]{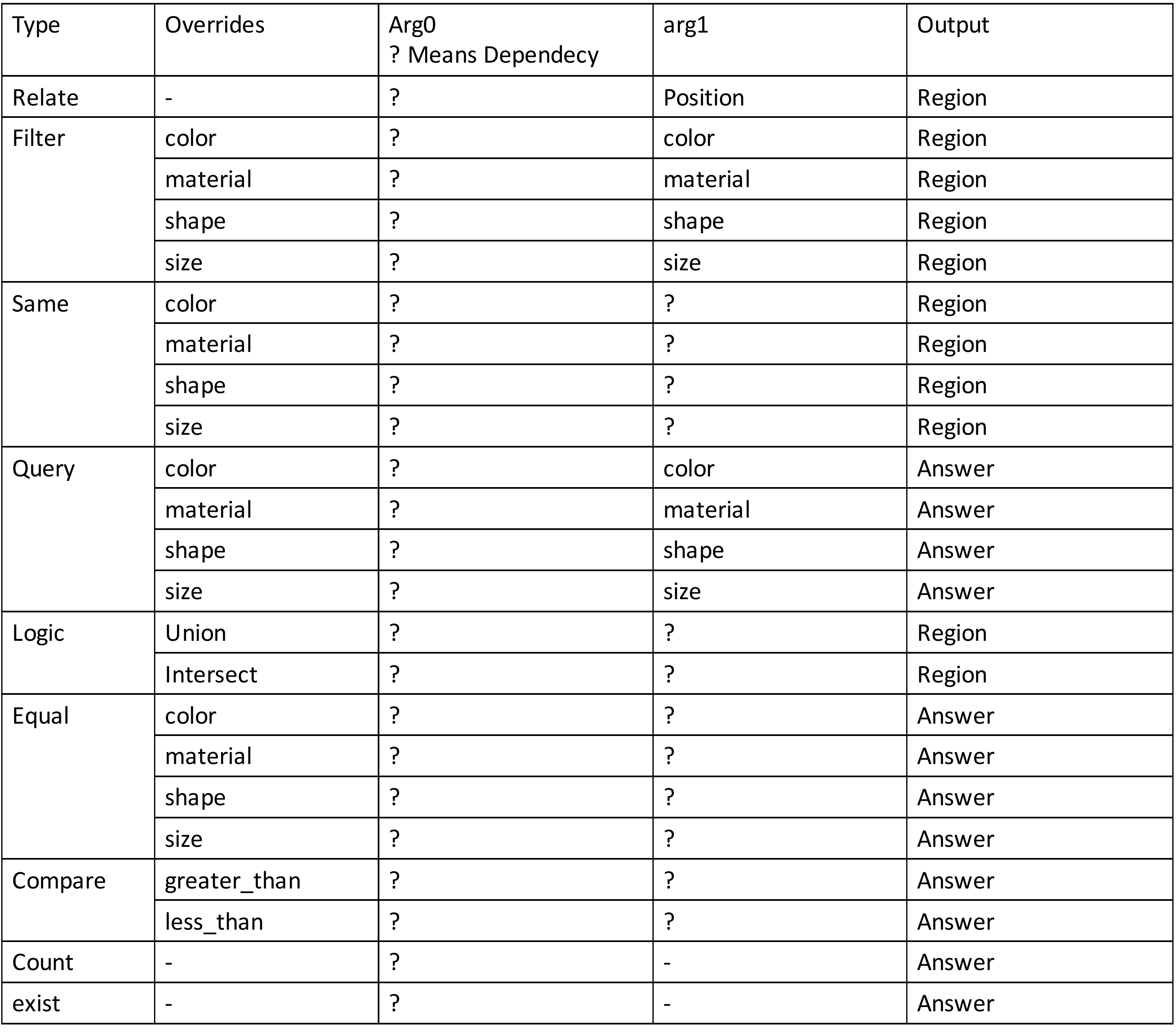}
    \caption{The function definitions and their corresponding outputs on CLEVR.}
    \label{fig:functions_clevr}
\end{figure}

\begin{figure}[!h]
    \centering
    \includegraphics[width=0.90\linewidth]{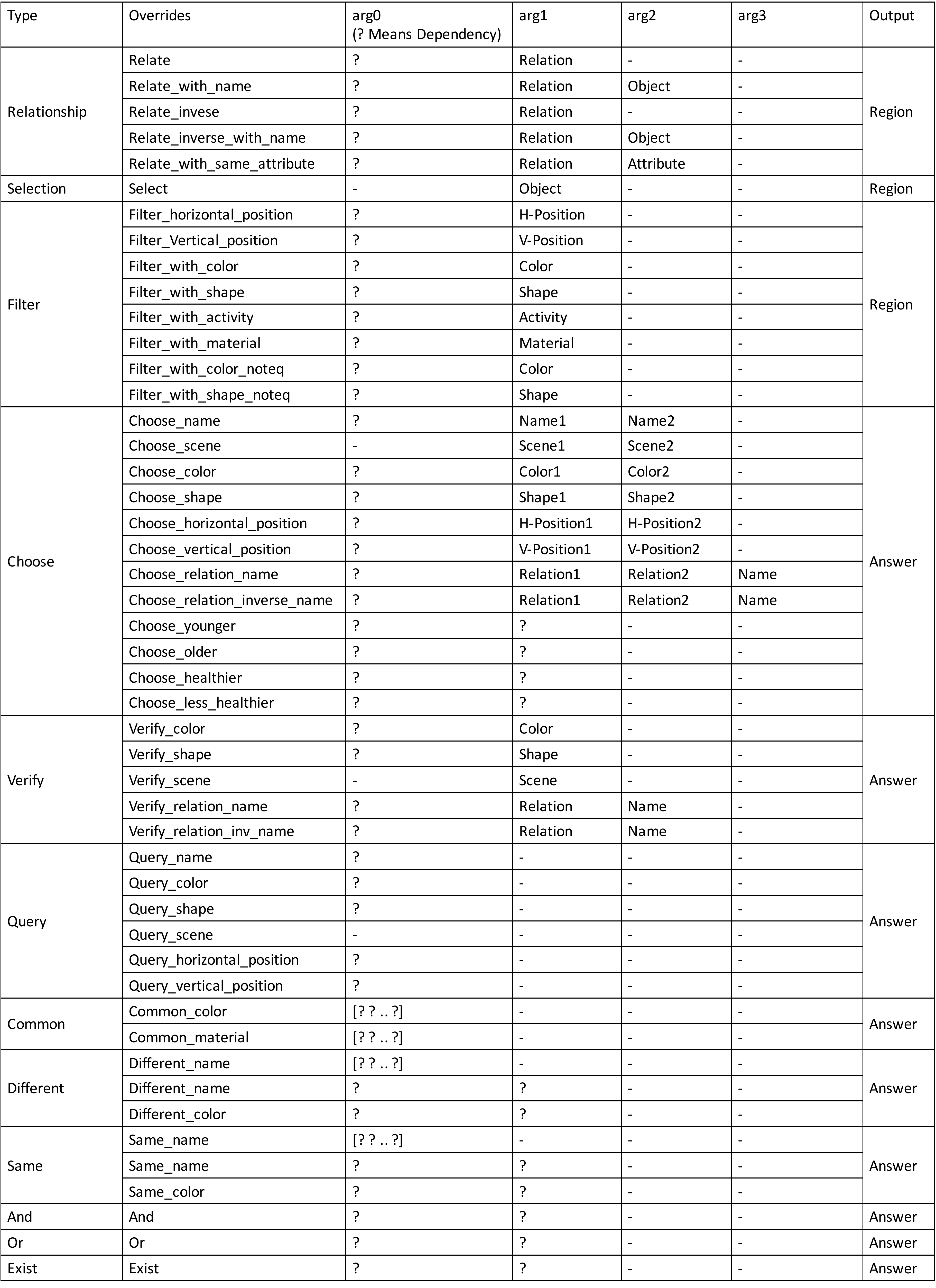}
    \caption{The function definitions and their corresponding outputs on GQA.}
    \label{fig:functions}
\end{figure}

The detailed function descriptions for CLEVR and GQA are provided in Figure~\ref{fig:functions_clevr} and Figure~\ref{fig:functions}, respectively.

\end{document}